\documentclass[10pt,twocolumn,letterpaper]{article}
\usepackage[T1]{fontenc}
\usepackage[utf8]{inputenc}
\usepackage[letterpaper,margin=0.75in,columnsep=0.28in]{geometry}
\usepackage{mathptmx}
\usepackage{amsmath,amssymb}
\usepackage{graphicx}
\usepackage{xcolor}
\definecolor{codeblue}{RGB}{0,84,166}
\usepackage{booktabs,multirow}
\usepackage{tabularx}
\usepackage{pifont}
\usepackage[round,authoryear]{natbib}
\usepackage[font=small,labelfont=bf]{caption}
\usepackage{newfloat}
\DeclareFloatingEnvironment[name=Algorithm,placement=tbp]{algorithm}
\usepackage{tikz}
\usetikzlibrary{arrows.meta,decorations.pathreplacing,calc,positioning,shapes.geometric}
\usepackage{microtype}
\usepackage{placeins}
\usepackage[hyphens]{url}
\usepackage[hidelinks]{hyperref}
\hypersetup{
  pdftitle={SyncHeC: Markerless Spatio-Temporal Hand-Eye Calibration via Differentiable Rendering and Continuous-Time Joint Optimization},
  pdfauthor={Zhengdong Hong},
  pdfsubject={Markerless spatial and temporal calibration for robot manipulation}
}
\newcommand{\SE}{\mathrm{SE}(3)}
\newcommand{\FK}{\mathrm{FK}}
\newcommand{\dt}{\Delta t}

\DeclareTextFontCommand{\texttt}{\fontencoding{OT1}\fontfamily{cmtt}\selectfont}
\title{SyncHeC: Markerless Spatio-Temporal Hand-Eye Calibration via\\
Differentiable Rendering and Continuous-Time Joint Optimization}
\author{Zhengdong Hong\\Zhejiang University}
\date{}
\begin{document}
\maketitle
\begin{abstract}
Markerless hand-eye calibration under continuous motion requires jointly resolving camera geometry and the time offset between images and joint states. We present SyncHeC (Synchronized Hand-Eye Calibration), a dense differentiable-rendering framework that estimates both from a single robot trajectory. Continuous-time forward kinematics makes latency differentiable, metric depth complements mask contours, and an offline planner combines spatial sensitivity with temporal excitation. Alternating updates support Eye-to-Hand, Eye-in-Hand, and multi-camera calibration. On a 7-DoF arm, one $6.5$\,s motion yields $0.21$\,cm translation error against a marker-board reference, $0.34$\,cm touch-test error, and $2.7$\,ms latency error relative to an $87$\,ms reference with $95\%$ uncertainty of $\pm1.5$\,ms. Against CM-Calib on matched recordings, translation and nominal latency errors improve by $3.7\times$ and $2.3\times$. Experiments span seven arms and four depth sensors. Compensating the recovered offset during policy training and deployment increases success by $8.1$ percentage points across six families and six tasks ($1{,}800$ matched pairs; paired $95\%$ confidence interval $[6.3,9.9]$ points).
\end{abstract}
\par\smallskip
\noindent\href{https://github.com/hongzhengdong/SyncHeC}{\textcolor{codeblue}{\textbf{Code: github.com/hongzhengdong/SyncHeC}}}\par

\section{Introduction}
\label{sec:introduction}
\begin{figure*}[t]
\centering
\includegraphics[width=0.92\linewidth]{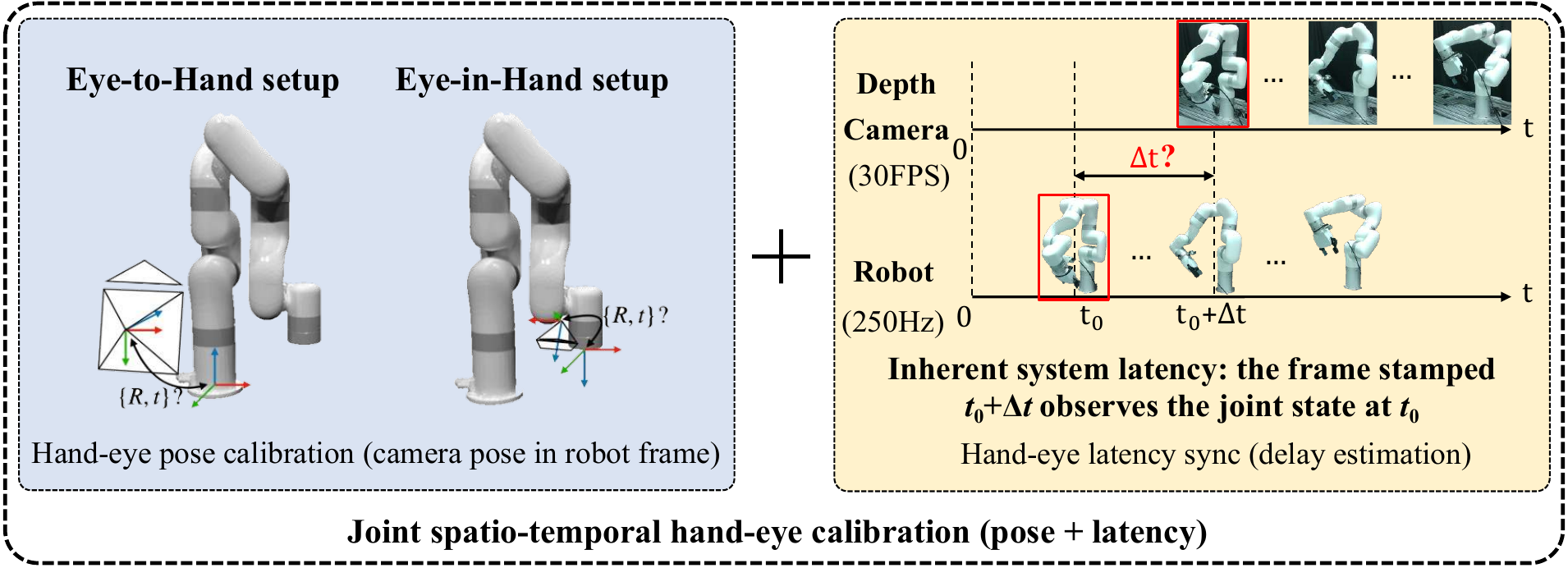}
\caption{Robotic hand-eye coordination has two coupled facets. \textbf{Left:} spatial pose calibration recovers the camera pose in the robot frame, in both Eye-to-Hand and Eye-in-Hand setups. \textbf{Right:} temporal calibration recovers the sensor-to-control latency $\Delta t$: a frame time-stamped at $t_0{+}\Delta t$ actually observes the joint state at $t_0$. We solve both jointly.}
\label{fig:teaser}
\end{figure*}

Visual robot control depends on two correspondences (Figure~\ref{fig:teaser}). Spatial calibration maps camera observations into the robot's kinematic frame~\citep{tsaiNewTechniqueFully1989,shiuCalibrationWristMounted1989,daniilidisHandEyeCalibrationUsing1999,ilonenRobustRobotcameraCalibration2011}. Temporal calibration associates each observation with the joint state it depicts~\citep{wuCorrespondenceMatchingTime2020}. Exposure, transport, processing, and clock offsets separate a frame's recorded timestamp from its physical observation time. During motion, this mismatch changes the robot geometry associated with an image and couples a timing error to the estimated extrinsics.

The same correspondence matters for learned control. Vision-language-action models~\citep{blackVisionLanguageActionFlow2025,kimOpenVLAOpenSourceVisionLanguageAction2024,liuRdt1bDiffusionFoundation2025} and 3D policies such as R3D~\citep{hongR3D2026} consume visual and proprioceptive information to predict actions. A calibration layer that aligns these streams provides geometrically and temporally consistent training tuples and deployment observations across policy architectures.

Marker-based methods and active-vision extensions estimate extrinsics from target observations~\citep{zhangActiveHandeyeCalibration2023}. Markerless methods use robot keypoints or direct pose regression~\citep{leeCameratorobotPoseEstimation2020,valassakisLearningEyeinhandCamera2022} or align a rendered robot model with image masks~\citep{chenEasyhecAccurateAutomatic2023,hongEasyHeCFullyAutomatic2024}. These approaches enable automatic spatial calibration from static poses. CM-Calib~\citep{wuCorrespondenceMatchingTime2020} jointly estimates pose and latency under continuous motion through target-derived correspondences. \textbf{SyncHeC} combines continuous-motion calibration with dense, markerless supervision (Table~\ref{tab:methods}).

\begin{table*}[t]
\centering
\small
\setlength{\tabcolsep}{4pt}
\caption{Capability comparison of representative hand-eye calibration paradigms. \emph{Marker-free}, \emph{EiH} (Eye-in-Hand support), \emph{Latency} (estimates $\Delta t$), \emph{Hetero.} (heterogeneous arms), \emph{Dyn.} (calibrates during continuous motion).}
\label{tab:methods}
\begin{tabular*}{\textwidth}{@{\extracolsep{\fill}}lccccc@{}}
\toprule
\textbf{Method} & \textbf{Marker-free} & \textbf{EiH} & \textbf{Latency} & \textbf{Hetero.} & \textbf{Dyn.}\\
\midrule
Marker-based \citep{tsaiNewTechniqueFully1989} & \ding{55} & \ding{51} & \ding{55} & \ding{51} & \ding{55}\\
Pose networks \citep{leeCameratorobotPoseEstimation2020,valassakisLearningEyeinhandCamera2022} & \ding{51} & \ding{51} & \ding{55} & \ding{55} & \ding{55}\\
EasyHeC \citep{chenEasyhecAccurateAutomatic2023} & \ding{51} & \ding{55} & \ding{55} & \ding{51} & \ding{55}\\
EasyHeC++ \citep{hongEasyHeCFullyAutomatic2024} & \ding{51} & \ding{51} & \ding{55} & \ding{51} & \ding{55}\\
CM-Calib \citep{wuCorrespondenceMatchingTime2020} & \ding{55} & \ding{51} & \ding{51} & \ding{51} & \ding{51}\\
\textbf{SyncHeC}     & \ding{51} & \ding{51} & \ding{51} & \ding{51} & \ding{51}\\
\bottomrule
\end{tabular*}
\end{table*}

SyncHeC integrates three components. First, interpolating the high-rate joint log creates a continuous-time kinematic model through which gradients reach the unknown latency. Second, metric RGB-D depth supplements mask contours with direct surface-distance constraints, strengthening optical-axis pose estimation. Third, a planned trajectory supplies spatially diverse views and sustained image motion for joint estimation. An alternating optimizer updates the temporal and spatial variables in turn.

SyncHeC supports both Eye-to-Hand (EtH) and Eye-in-Hand (EiH) cameras, and extends to multiple cameras through a common kinematic model. One short acquisition replaces repeated stop-and-go motion. We evaluate spatial accuracy, timing accuracy, conditioning, depth processing, and hardware transfer, then test the value of the recovered offset through matched downstream manipulation trials.

\section{Related Work}
\label{sec:related}
\paragraph{Spatial hand-eye calibration.}
Classical solvers estimate $\mathbf{AX}=\mathbf{XB}$ or $\mathbf{AX}=\mathbf{YB}$ from calibration targets and forward kinematics~\citep{tsaiNewTechniqueFully1989,shiuCalibrationWristMounted1989,daniilidisHandEyeCalibrationUsing1999,ilonenRobustRobotcameraCalibration2011}. Active approaches select informative viewpoints~\citep{zhangActiveHandeyeCalibration2023,yangNextbestviewSelectionRobot2023}. Markerless networks predict keypoints or camera poses~\citep{leeCameratorobotPoseEstimation2020,valassakisLearningEyeinhandCamera2022}. EasyHeC and EasyHeC++ optimize rendered robot silhouettes, with pretrained image models supporting automatic initialization~\citep{chenEasyhecAccurateAutomatic2023,hongEasyHeCFullyAutomatic2024}. Hydra registers RGB-D observations to the robot model~\citep{huberHydraMarkerFreeRGBD2025}. Our dense spatial objective adds a differentiable timing variable to robot-model alignment.

\paragraph{Differentiable rendering and dynamic reconstruction.}
Differentiable rasterization propagates image residuals to scene geometry~\citep{loperOpenDRApproximateDifferentiable2014,katoDifferentiableRenderingSurvey2020,laineModularPrimitivesHighperformance2020}, supporting inverse rendering~\citep{kayaUncalibratedNeuralInverse2021,kerbl3dGaussianSplatting2023} and object pose estimation~\citep{wenFoundationPoseUnified6D2024}. DRRobot differentiates a kinematics-aware robot appearance model with respect to joint controls~\citep{liuDifferentiableRobotRendering2025}; RobotArena$\infty$ uses differentiable rendering for camera-pose estimation in real-to-sim reconstruction~\citep{jangirRobotArena2026}. Dynamic free-viewpoint reconstruction with a flying camera also couples camera calibration and articulated motion estimation~\citep{hongFreeViewpoint2025}. Here, recorded robot kinematics provide a measured trajectory, and shifting its time argument makes sensor latency directly observable through rendered masks and depth.

\paragraph{Temporal calibration and informative motion.}
Camera--IMU timing is estimated jointly with motion in visual-inertial systems~\citep{liOnlineTemporalCalibration2014,qinVinsmonoRobustVersatile2018}, including continuous-time batch formulations~\citep{furgaleContinuoustimeBatchEstimation2012}. Related methods address targetless LiDAR--IMU and LiDAR--camera calibration~\citep{lvTargetlessCalibrationLiDARIMU2020,liTwostepLiDARCamera2023}. For arms, CM-Calib uses correspondence matching~\citep{wuCorrespondenceMatchingTime2020}, while laser-vision methods exploit structured features~\citep{yangSpatiotemporalCalibrationLaser2025}. SyncHeC estimates latency through dense mask and depth observations of the robot. Our planner combines spatial sensitivity with temporal excitation; this extends informative-view selection~\citep{jiangFisherRF2024,jinNeuNBV2023} to a timed robot trajectory.

\paragraph{Geometry and dynamics for manipulation.}
Several complementary approaches organize manipulation around geometric or motion representations. R3D studies 3D policy encoders and training~\citep{hongR3D2026}; LAMP lifts image-editing cues into 3D inter-object transformations~\citep{wangLAMP2026}. AdaDexGrasp uses single human demonstrations to guide grasp-skill learning~\citep{shiAdaptiveDexterous2025}, and object-centric motion priors transfer predicted human--object states to robot learning~\citep{hongObjectCentric2026}. Particle-based world models learn interaction dynamics from human data and generate reference trajectories through model predictive control~\citep{hongParticleWorldModel2025}. Our downstream study examines the observation-to-joint alignment used by image policies, point-cloud policies, and a particle-based dynamics controller.

\section{Method}
\label{sec:method}
\begin{figure*}[t]
\centering
\includegraphics[width=0.98\linewidth]{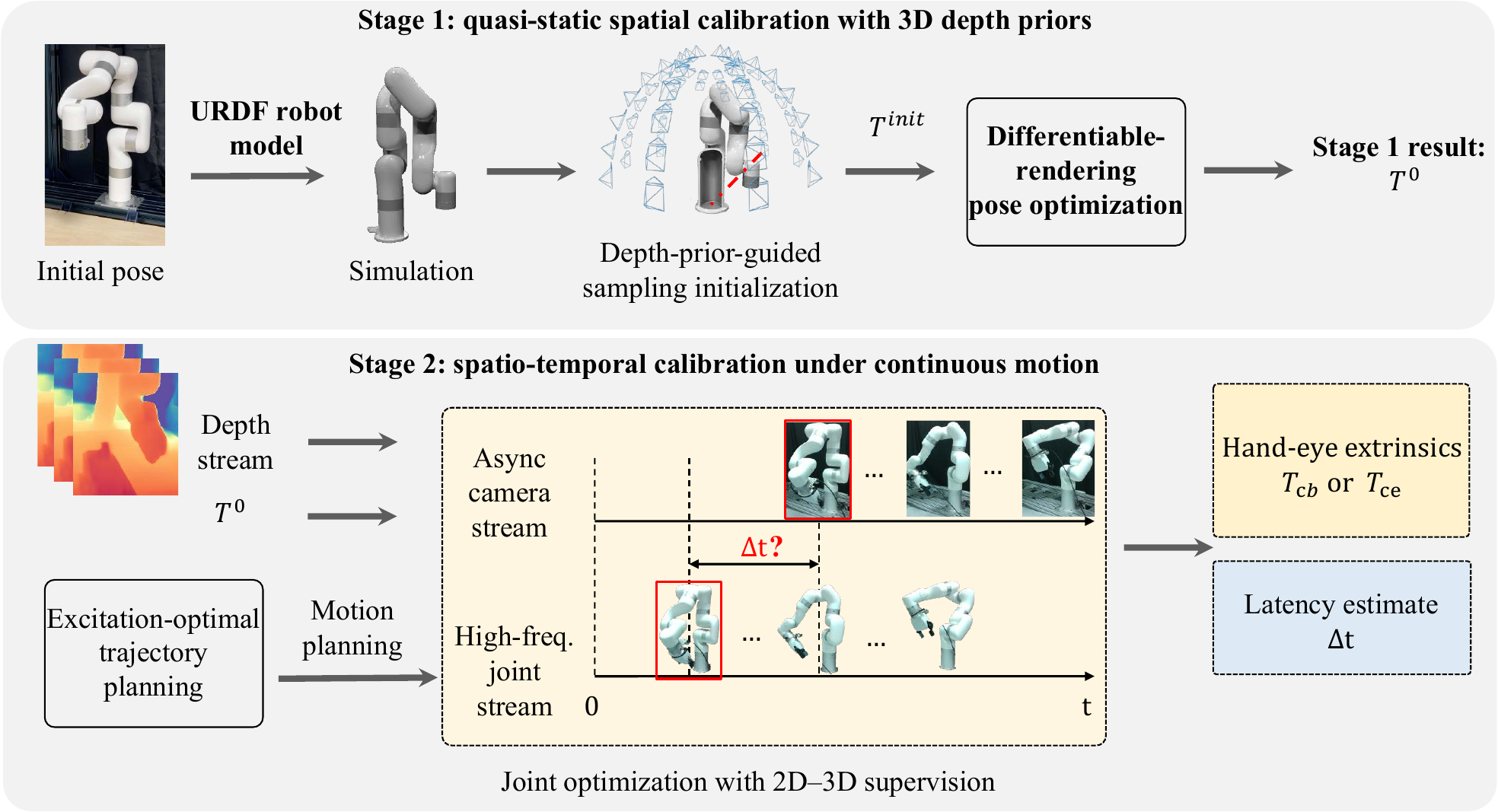}
\caption{Overview. \textbf{Stage 1} imports the URDF into a simulator, sizes an adaptive sampling sphere from the first real depth frame, selects an initial pose by a 2D--3D fitness, and refines it by differentiable rendering to obtain $\mathbf{T}^0$. \textbf{Stage 2} executes one offline-planned continuous trajectory, parameterizes the high-frequency joint stream as continuous-time forward kinematics, and jointly optimizes the extrinsics and latency $\Delta t$ from a 2D--3D differentiable-rendering loss.}
\label{fig:pipeline}
\end{figure*}

SyncHeC couples continuous-time robot kinematics with dense rendering losses to estimate an extrinsic transform and an image--joint offset. Figure~\ref{fig:pipeline} summarizes spatial initialization, trajectory acquisition, and alternating optimization.
\subsection{Observation Model}
\label{sec:observation}
Let $\mathbf{T}_{ab}$ map coordinates from frame $b$ to frame $a$. For EtH we estimate the base-to-camera transform $\mathbf{T}_{cb}\in\SE$; for EiH we estimate the flange-to-camera transform $\mathbf{T}_{ce}\in\SE$. Their inverses map camera coordinates into the corresponding robot frame. RGB-D frames have timestamps $\{t_i^c\}_{i=1}^N$, and joint measurements form a denser log $\mathcal{Q}=\{(t_j^r,\mathbf{q}_j)\}_{j=1}^M$. The frame stamped $t_i^c$ observes the joint state at
\begin{equation}
  \tau_i=t_i^c-\dt.
  \label{eq:time}
\end{equation}
The constant offset $\dt$ includes sensing-to-availability delay and the difference between clock origins. It specifies image--joint pairing during acquisition, training, and deployment.

For link $l$, $\FK^{(l)}(\mathbf{q})$ maps link coordinates into the robot base, and $\FK^{(e)}$ denotes flange kinematics~\citep{carpentierPinocchioLibraryFast2019}. The two camera models are
\begin{align}
\mathbf{T}_{c l}^{\mathrm{EtH}}(\tau_i)
  &=\mathbf{T}_{cb}\FK^{(l)}(\mathbf{q}(\tau_i)), \label{eq:eth}\\
\mathbf{T}_{c l}^{\mathrm{EiH}}(\tau_i)
  &=\mathbf{T}_{ce}
  [\FK^{(e)}(\mathbf{q}(\tau_i))]^{-1}
  \FK^{(l)}(\mathbf{q}(\tau_i)). \label{eq:eih}
\end{align}
Given link meshes, camera intrinsics, and a differentiable renderer, these transforms produce a mask and depth image for each observation. We parameterize the static extrinsic by $\xi\in\mathfrak{se}(3)$ and estimate $(\xi,\dt)$ jointly.

\subsection{Metric Depth and Spatial Initialization}
\label{sec:depth}
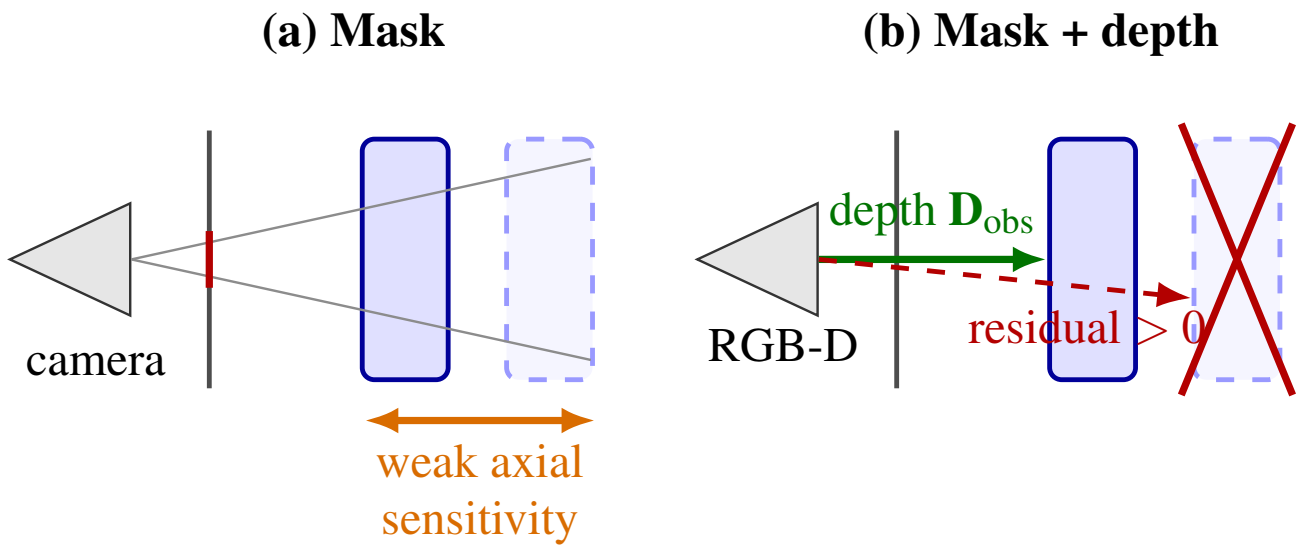
\begin{figure}[t]
\centering
\resizebox{\linewidth}{!}{%
\begin{tikzpicture}[
    x=0.60cm, y=0.75cm,
    every node/.style = {font=\small},
    cam/.style   = {isosceles triangle, isosceles triangle apex angle=50, draw=black!80,
                     fill=black!10, minimum height=0.7cm, shape border rotate=180, inner sep=1pt},
    arm/.style   = {rectangle, rounded corners=2pt, draw=blue!60!black, thick, fill=blue!12,
                     minimum width=0.5cm, minimum height=1.4cm},
    armg/.style  = {rectangle, rounded corners=2pt, draw=blue!40, dashed, thick, fill=blue!4,
                     minimum width=0.5cm, minimum height=1.4cm},
    ray/.style   = {black!45, thin},
    lbl/.style   = {font=\footnotesize, align=center},
]
\begin{scope}
\node[cam] (c1) at (0,0) {};
\node[lbl, below=5pt of c1] {camera};
\draw[black!70, thick] (1.1,-1.0) -- (1.1,1.0);
\node[arm]  (a1) at (3.0,0) {};
\node[armg] (a2) at (4.4,0) {};
\draw[ray] (c1.east) -- (4.8,0.78);
\draw[ray] (c1.east) -- (4.8,-0.78);
\draw[red!70!black, very thick] (1.1,0.22) -- (1.1,-0.22);
\draw[{Latex[length=2mm]}-{Latex[length=2mm]}, orange!85!black, very thick]
      (2.6,-1.25) -- (4.85,-1.25) node[midway, below, lbl, orange!85!black]{weak axial\\sensitivity};
\node[lbl, font=\footnotesize\bfseries] at (2.5,1.75) {(a) Mask};
\end{scope}
\begin{scope}[xshift=4.0cm]
\node[cam] (c2) at (0,0) {};
\node[lbl, below=1pt of c2] {RGB-D};
\draw[black!70, thick] (1.1,-1.0) -- (1.1,1.0);
\node[arm]  (b1) at (3.0,0) {};
\node[armg] (b2) at (4.4,0) {};
\draw[-{Latex[length=2mm]}, green!45!black, very thick] (c2.east) -- (b1.west)
      node[midway, above, lbl, green!45!black]{depth $\mathbf{D}_{\mathrm{obs}}$};
\draw[-{Latex[length=2mm]}, red!70!black, thick, dashed] (c2.east) -- ($(b2.west)+(0,-0.3)$)
      node[pos=0.72, below, lbl, red!70!black]{residual $>0$};
\draw[red!70!black, very thick] ($(b2.north east)+(0.1,0.1)$) -- ($(b2.south west)+(-0.1,-0.1)$);
\draw[red!70!black, very thick] ($(b2.north west)+(-0.1,0.1)$) -- ($(b2.south east)+(0.1,-0.1)$);
\node[lbl, font=\footnotesize\bfseries] at (2.5,1.75) {(b) Mask + depth};
\end{scope}
\end{tikzpicture}%
}
\caption{Complementary geometric constraints. Solid and dashed shapes show reference and axially shifted geometry. (a)~Mask contours constrain axial translation through perspective, with weak sensitivity in distant or narrow projections. (b)~Metric depth adds a direct surface-distance residual.}
\label{fig:ambiguity}
\end{figure}

\paragraph{Axial conditioning.}
Known-scale perspective geometry carries an optical-axis constraint: for $u=fX/Z$, $\partial u/\partial Z=-fX/Z^2$. Its sensitivity becomes small for distant or narrow projections, and mask contours provide limited support along this direction. Metric depth adds direct surface-distance residuals over the visible robot, strengthening axial conditioning (Figure~\ref{fig:ambiguity}).

\paragraph{Sensor-aligned depth.}
We use the metric Hypersim checkpoint of Depth Anything V2 (ViT-L)~\citep{yangDepthAnythingV22024}. Let $D_{\rm raw}$ be sensor depth and $D_{\rm pred}$ the predicted depth, both registered to the RGB grid. On pixels where both are valid, their median ratio sets a scale $s$. We retain valid sensor measurements and fill holes with $sD_{\rm pred}$; nonpositive and nonfinite values are excluded. The resulting metric map $D_{\rm obs}$ and validity mask are shared by initialization and dynamic calibration. Appendix~\ref{app:depth} specifies the fusion and supervision region.

\paragraph{Stage 1: spatial anchoring.}
A single static RGB-D frame initializes the dynamic pipeline. We load the robot URDF into SAPIEN~\citep{xiangSAPIENSimulAtedPartBased2020}, segment its base link with SAM2~\citep{raviSAM2Segment2025}, and take the median valid depth as a surface-distance anchor $d_{\rm surf}$. A side-profile correction estimates the base-center distance $L$ from this surface anchor (Appendix~\ref{app:depth}). The camera-sampling sphere has radius $R_{\rm adaptive}=L+\Delta r$, with margin $\Delta r=0.10$\,m.

For each candidate extrinsic $\mathbf{T}_s$, we render its mask and depth and evaluate
\begin{equation}
S(\mathbf{T}_s)=\operatorname{IoU}(\mathbf{T}_s)
-\frac{\lambda_{\rm init}}{|\Omega_s|}
\sum_{p\in\Omega_s}|D_{\rm rend}^{(p)}(\mathbf{T}_s)-D_{\rm obs}^{(p)}|,
\label{eq:fitness}
\end{equation}
where $\Omega_s$ contains valid overlapping robot pixels. Mask-based rendering refinement of the best candidate produces $\mathbf{T}^0$. The five-view static benchmark additionally refines the extrinsic using the remaining static views.

\subsection{Continuous-Time Joint Calibration}
\label{sec:joint}
\paragraph{Differentiable latency.}
For $\tau_i\in[t_k^r,t_{k+1}^r)$, the synchronized joint state and its temporal derivative are
\begin{align}
\mathbf{q}(\tau_i)
 &=\mathbf{q}_k+
 \frac{t_i^c-\dt-t_k^r}{t_{k+1}^r-t_k^r}
 (\mathbf{q}_{k+1}-\mathbf{q}_k), \label{eq:interp}\\
\frac{\partial\mathbf{q}(\tau_i)}{\partial\dt}
 &=-\frac{\mathbf{q}_{k+1}-\mathbf{q}_k}{t_{k+1}^r-t_k^r}.
 \label{eq:timejac}
\end{align}
The active interval is recomputed as $\dt$ changes. With a $250$\,Hz log, adjacent samples are $4$\,ms apart; bounded joint acceleration gives an $O(h^2)$ joint-value interpolation error. Substitution into Equations~\eqref{eq:eth}--\eqref{eq:eih} propagates time gradients through forward kinematics and rendering.

\paragraph{Dense 2D--3D objective.}
SAM2 / Grounded-SAM provide observation masks~\citep{kirillovSegmentAnything2023,raviSAM2Segment2025,renGroundedSAMAssembling2024}. For frame $i$, the rasterizer outputs a soft mask $M_i^{\rm rend}$ and metric depth $D_i^{\rm rend}$. We minimize
\begin{align}
\mathcal{L}_{m,i}
 &=\frac{1}{HW}\sum_p(M_i^{{\rm rend},(p)}-M_i^{{\rm obs},(p)})^2,\\
\mathcal{L}_{d,i}
 &=\frac{1}{|\Omega_i|}\sum_{p\in\Omega_i}
 |D_i^{{\rm rend},(p)}-D_i^{{\rm obs},(p)}|,\\
\mathcal{L}(\xi,\dt)
 &=\frac1N\sum_{i=1}^N
 (\mathcal{L}_{m,i}+\lambda_D\mathcal{L}_{d,i})
 +\lambda_{\rm reg}\dt^2.
\label{eq:loss}
\end{align}
Here $\Omega_i$ intersects the observed robot mask, rendered mask $>0.5$, and valid observed depth. The depth term is zero for an empty intersection. Depth is measured in meters and $\dt$ in seconds.

\paragraph{Joint observability.}
Stack the rendered observations across frames and differentiate with respect to the six pose parameters and time:
\begin{equation}
\mathbf{J}=[\,\mathbf{J}_{\xi}\ \mathbf{j}_t\,],
\qquad
\mathbf{j}_t=-\mathbf{J}_{\pi}\dot{\mathbf q},
\label{eq:jointjac}
\end{equation}
where the expression for $\mathbf j_t$ is stacked framewise and $\mathbf{J}_{\pi}$ includes the appropriate EtH or EiH kinematic chain. Local first-order identifiability requires $\operatorname{rank}(\mathbf{J})=7$~\citep{hornMatrixAnalysis2012}. A temporal change becomes indistinguishable from a pose change when its stacked column lies in the span of the six pose columns. Exciting multiple configurations and velocities helps separate these effects. This criterion depends on the full projected motion across the sequence.

\paragraph{Alternating updates.}
Starting from $\mathbf{T}^0$ and $\dt=0$, we alternate a temporal block with a spatial block. The temporal block fixes pose and updates $\dt$ using frames with $\|\dot{\mathbf q}(t_i^c-\dt)\|_2>0.05$\,rad/s. The spatial block resynchronizes all selected frames and updates pose. Both use Adam with persistent optimizer moments. Each block performs $15$ passes over mini-batches of four frames. In the dynamic benchmark, all $60$ selected frames pass the temporal gate in every cycle, giving $225$ actual Adam updates per variable per cycle. Convergence takes $3$--$5$ cycles, or $675$--$1{,}125$ temporal updates. Appendix~\ref{app:optimizer} gives the algorithm and a representative trace.

\subsection{Excitation-Based Trajectory Planning}
\label{sec:planning}
We optimize a short timed trajectory using spatial and temporal rendering sensitivities. For a candidate camera set $\mathbf T_{\rm cand}$, spatial excitation is
\begin{equation}
V_{\rm space}(\mathbf q)
=\sum_p\operatorname{Var}_{\mathbf T\in\mathbf T_{\rm cand}}
\bigl(\pi^{(p)}(\mathbf T,\mathbf q)\bigr).
\label{eq:vspace}
\end{equation}
The rendered image $\pi$ uses the appropriate observation model in Section~\ref{sec:observation}. Temporal excitation is
\begin{equation}
V_{\rm time}(\mathbf q,\dot{\mathbf q})
=\|\mathbf J_\pi(\mathbf q)\dot{\mathbf q}\|_F.
\label{eq:vtime}
\end{equation}
These terms measure pose sensitivity and the magnitude of the rendered observation's temporal derivative. For Gaussian observation noise, squared sensitivity determines a Fisher-information block; our unsquared temporal score rewards motion over the sampled trajectory. Appendix~\ref{app:conditioning} derives this relationship and describes the joint conditioning diagnostic.

RRT*~\citep{karamanSamplingbasedAlgorithmsOptimal2011} supplies a collision-free geometric seed, followed by optimization of a timed $K$-knot B-spline:
\begin{equation}
\mathbf Q^*=\arg\max_{\mathbf Q}
\sum_{k=1}^K\left[
\alpha V_{\rm space}(\mathbf q_k)+
\beta V_{\rm time}(\mathbf q_k,\dot{\mathbf q}_k)
\right].
\label{eq:planner}
\end{equation}
The refinement enforces joint limits, collision avoidance, and base-link visibility. Planning uses the Stage-1 extrinsic. We normalize the two scores by fixed initial-reference scales $s_{\rm space}=2000$ and $s_{\rm time}=400$, giving $\alpha=1/s_{\rm space}=5\times10^{-4}$ and $\beta=\gamma/s_{\rm time}=2.5\times10^{-3}$. The relative weight $\gamma=1$ is selected on the simulation split and held fixed for real experiments.

\subsection{Multi-Camera Calibration}
\label{sec:multimethod}
For several cameras, we sum their normalized rendering objectives, assign each camera its own extrinsic, and evaluate their observations through the shared joint log. The model admits a separate $\dt_j$ per camera. When paired timing measurements establish equivalence within a specified tolerance, a shared-offset variant imposes $\dt_1=\dt_2$. The temporal block then accumulates evidence from both views, while each spatial block updates its own extrinsic. Our dual-camera experiment uses a $\pm1$\,ms equivalence criterion verified by an independent paired timing reference.

\section{Experiments}
\label{sec:experiments}
\subsection{Setup and Evaluation}
\label{sec:setup}
The main platform is a 7-DoF xArm7 with environment-mounted EtH and wrist-mounted EiH RealSense D435i cameras. Both stream $848\times480$ RGB-D at $30$\,Hz, and the controller records joints at $250$\,Hz. Rendering and optimization use Nvdiffrast~\citep{laineModularPrimitivesHighperformance2020} on an RTX 4090 with an i9-13900K. The planned $25$-knot trajectory lasts approximately $6.5$\,s. We retain the $60$ highest-flow frames from approximately $195$ recorded frames.

\paragraph{Optimization settings.}
We set $\lambda_D=0.1$, $\lambda_{\rm reg}=0.5$, and $\lambda_{\rm init}=1.0$. Adam~\citep{kingmaAdamMethodStochastic2015} uses learning rates $10^{-3}$ for rotation, $5\times10^{-4}$ for translation, and $2\times10^{-4}$ for time, with moments $(0.9,0.999)$ and numerical epsilon $10^{-8}$. Each temporal and spatial block uses $15$ passes, batch size four, and persistent moments. At most $15$ cycles give $3{,}375$ temporal updates. We stop when cycle-to-cycle changes are below $0.05$\,mm in translation, $0.01^\circ$ in rotation, and $0.5$\,ms in time. Loss weights are selected on the simulation split from $\lambda_D\in\{0.05,0.1,0.2,0.5\}$ and $\lambda_{\rm reg}\in\{0.1,0.5,1\}$, then held fixed across real experiments. Varying $\lambda_D$ within $[0.05,0.2]$ changes translation error by less than $0.03$\,cm.

\paragraph{Independent references.}
A signal-generator, LED, and oscilloscope protocol relates image exposure to the logged physical onset of an arm micro-motion. Exposure-phase, clock-mapping, and physical-onset measurements bound residual biases by $0.25$, $0.10$, and $0.20$\,ms, respectively; the phase budget includes RGB--depth timestamp skew. Twenty repetitions give an $87$\,ms nominal reference with approximately $2$\,ms single-repeat spread and a $0.95$\,ms confidence half-width for the mean. Combining that interval with the three residual bounds gives $95\%$ reference uncertainty $U_{95}=1.5$\,ms (Appendix~\ref{app:timing}). Spatial references are obtained independently from repeated marker-board measurements and a workspace touch test.

\paragraph{Metrics and comparisons.}
We report translation error $E_t$ in cm, geodesic rotation error $E_r$ in degrees, and nominal latency error $E_{\rm lat}=|\dt^*-\dt_{\rm ref}|$ in ms. Marker-board pose error measures extrinsic agreement; the touch test additionally includes probe-offset and robot-positioning errors. Spatial reference agreement is approximately $1$\,mm.

Static baseline results in Tables~\ref{tab:sim}--\ref{tab:real_static} follow the simulation and touch protocols reported by EasyHeC++~\citep{hongEasyHeCFullyAutomatic2024}; SyncHeC reruns those protocols with $\dt=0$. We also run Hydra under the matched static touch protocol. On dynamic data, CM-Calib and SL-Calib use the same planned trajectory, recordings, and joint logs as SyncHeC, with initialization matched where applicable. CM-Calib receives target-derived poses and forward kinematics. SL-Calib is our straight-line-motion implementation of the correspondence-based approach. Appendix~\ref{app:comparisons} records comparison provenance and controls.

\subsection{Static Spatial Accuracy}
In simulation (Table~\ref{tab:sim}), our translation errors are $0.105$\,cm for EtH and $0.086$\,cm for EiH, compared with $0.135$ and $0.112$\,cm for EasyHeC++. On the real touch test (Table~\ref{tab:real_static}), EtH performance is comparable across SyncHeC, EasyHeC++, and Hydra at the reference resolution. EiH error decreases from $0.31$\,cm for EasyHeC++ and $0.28$\,cm for Hydra to $0.16$\,cm. Metric depth and the surface-to-base correction are particularly useful in the constrained wrist-camera view.

\begin{table}[t]
\centering
\small
\setlength{\tabcolsep}{4pt}
\caption{Simulation calibration accuracy. Top: Eye-to-Hand. Bottom: Eye-in-Hand. \#v: number of views. Baseline values are reproduced from the EasyHeC++ evaluation~\citep{hongEasyHeCFullyAutomatic2024}.}
\label{tab:sim}
\begin{tabularx}{\columnwidth}{@{}>{\raggedright\arraybackslash}Xccc@{}}
\toprule
\textbf{Method} & \textbf{\#v} & $E_r$ ($^\circ$)$\downarrow$ & $E_t$ (cm)$\downarrow$\\
\midrule
\multicolumn{4}{l}{\emph{Eye-to-Hand}}\\
Marker-based~\citep{tsaiNewTechniqueFully1989} & 20 & 0.870 & 2.000\\
DREAM~\citep{leeCameratorobotPoseEstimation2020} & 5 & 0.704 & 0.303\\
EasyHeC~\citep{chenEasyhecAccurateAutomatic2023} & 5 & 0.081 & 0.206\\
EasyHeC++~\citep{hongEasyHeCFullyAutomatic2024} & 5 & 0.045 & 0.135\\
\textbf{SyncHeC}\,($\Delta t{=}0$) & 5 & \textbf{0.038} & \textbf{0.105}\\
\midrule
\multicolumn{4}{l}{\emph{Eye-in-Hand}}\\
\citet{zhangActiveHandeyeCalibration2023} & 10 & 0.148 & 0.315\\
\citet{valassakisLearningEyeinhandCamera2022} & 1 & 4.400 & 1.340\\
EasyHeC++~\citep{hongEasyHeCFullyAutomatic2024} & 5 & 0.117 & 0.112\\
\textbf{SyncHeC}\,($\Delta t{=}0$) & 5 & \textbf{0.095} & \textbf{0.086}\\
\bottomrule
\end{tabularx}
\end{table}

\begin{table}[t]
\centering
\small
\caption{Real static touch-test error (cm), averaged over ten poses. Hydra and SyncHeC are evaluated here; the other results are reproduced from \citet{hongEasyHeCFullyAutomatic2024}. Protocols are specified in Appendix~\ref{app:comparisons}.}
\label{tab:real_static}
\begin{tabularx}{\columnwidth}{@{}>{\raggedright\arraybackslash}Xcc@{}}
\toprule
\textbf{Method} & \textbf{EtH} & \textbf{EiH}\\
\midrule
DREAM~\citep{leeCameratorobotPoseEstimation2020} & 1.50 & ---\\
\citet{zhangActiveHandeyeCalibration2023} & --- & 1.35\\
\citet{valassakisLearningEyeinhandCamera2022} & --- & 4.30\\
EasyHeC~\citep{chenEasyhecAccurateAutomatic2023} & 0.40 & ---\\
EasyHeC++~\citep{hongEasyHeCFullyAutomatic2024} & 0.30 & 0.31\\
Hydra~\citep{huberHydraMarkerFreeRGBD2025} & 0.35 & 0.28\\
\textbf{SyncHeC} ($\Delta t=0$) & \textbf{0.28} & \textbf{0.16}\\
\bottomrule
\end{tabularx}
\end{table}

\subsection{Dynamic Spatio-Temporal Accuracy}
\label{sec:dynamic}
Table~\ref{tab:dynamic} evaluates continuous-motion calibration. Across five matched recordings, SyncHeC reaches $0.21\pm0.04$\,cm / $0.15\pm0.05^\circ$ / $2.7\pm0.6$\,ms for EtH, and $0.24\pm0.05$\,cm / $0.18\pm0.07^\circ$ / $5.4\pm0.9$\,ms for EiH. CM-Calib gives $0.78\pm0.08$ and $0.95\pm0.10$\,cm translation error, with latency errors $6.2\pm0.8$ and $8.5\pm1.1$\,ms. The EtH ratios are $3.7\times$ for translation and $2.3\times$ for nominal latency error.

The corresponding mean offset estimates are $84.3$/$81.6$\,ms for SyncHeC and $80.8$/$78.5$\,ms for CM-Calib (EtH/EiH). Our latency error remains smaller throughout the timing reference's $\pm1.5$\,ms uncertainty interval. For EtH, the CM-Calib-to-SyncHeC error ratio over this interval is $[1.83,3.92]$ (Appendix~\ref{app:timing}).

EasyHeC and EasyHeC++ applied directly to the moving sequence associate each frame with the unshifted joint state, giving EtH errors of $4.85$ and $3.92$\,cm. A fixed-reference timing control synchronizes the inputs at $87$\,ms before spatial optimization: EasyHeC++ then reaches $0.43$/$0.48$\,cm and the SyncHeC spatial objective reaches $0.20$/$0.23$\,cm (EtH/EiH). Joint estimation reaches essentially the same spatial accuracy as the fixed-reference version of SyncHeC. This control separates the benefit of temporal alignment from that of dense metric-depth supervision.

\begin{table}[t]
\centering
\small
\setlength{\tabcolsep}{4pt}
\caption{Continuous-motion calibration against the $87$\,ms reference ($U_{95}=1.5$\,ms). CM-Calib and ours use five matched recordings; SDs are shown for CM translation/latency and all our metrics.}
\label{tab:dynamic}
\setlength{\tabcolsep}{3pt}
\begin{tabularx}{\columnwidth}{@{}>{\raggedright\arraybackslash}Xccc@{}}
\toprule
\textbf{Method} & $E_t$ (cm)$\downarrow$ & $E_r$ ($^\circ$)$\downarrow$ & $E_{\text{lat}}$ (ms)$\downarrow$\\
\midrule
\multicolumn{4}{l}{\emph{Eye-to-Hand}}\\
SL-Calib$^\ddagger$ & 1.15 & 0.92 & 11.5\\
CM-Calib~\citep{wuCorrespondenceMatchingTime2020} & 0.78\,{\scriptsize$\pm$0.08} & 0.55 & 6.2\,{\scriptsize$\pm$0.8}\\
EasyHeC$^\dagger$~\citep{chenEasyhecAccurateAutomatic2023} & 4.85 & 3.42 & ---\\
EasyHeC++$^\dagger$~\citep{hongEasyHeCFullyAutomatic2024} & 3.92 & 2.75 & ---\\
\textbf{SyncHeC} & \textbf{0.21}\,{\scriptsize$\pm$0.04} & \textbf{0.15}\,{\scriptsize$\pm$0.05} & \textbf{2.7}\,{\scriptsize$\pm$0.6}\\
\midrule
\multicolumn{4}{l}{\emph{Eye-in-Hand}}\\
SL-Calib$^\ddagger$ & 1.34 & 1.15 & 14.8\\
CM-Calib~\citep{wuCorrespondenceMatchingTime2020} & 0.95\,{\scriptsize$\pm$0.10} & 0.72 & 8.5\,{\scriptsize$\pm$1.1}\\
EasyHeC$^\dagger$~\citep{chenEasyhecAccurateAutomatic2023} & 5.68 & 4.15 & ---\\
EasyHeC++$^\dagger$~\citep{hongEasyHeCFullyAutomatic2024} & 4.55 & 3.38 & ---\\
\textbf{SyncHeC} & \textbf{0.24}\,{\scriptsize$\pm$0.05} & \textbf{0.18}\,{\scriptsize$\pm$0.07} & \textbf{5.4}\,{\scriptsize$\pm$0.9}\\
\bottomrule
\end{tabularx}
\par\vspace{2pt}
{\footnotesize\raggedright $^\dagger$Spatial-only methods applied to the moving sequence with unshifted joint timestamps. Their fixed-reference control is reported in the text.\par}
{\footnotesize\raggedright $^\ddagger$SL-Calib is our straight-line-motion variant of CM-Calib~\citep{wuCorrespondenceMatchingTime2020}.\par}
\end{table}

\paragraph{Optimization behavior.}
A representative EtH run reaches $84.3$\,ms from zero in $900$ temporal updates (Appendix~\ref{app:optimizer}). Its data and regularizer gradients are approximately $-0.0843$ and $+0.0843$\,s$^{-1}$. Continuing alternation for $900$ additional temporal updates with stopping disabled changes the final offset by less than $0.2$\,ms, characterizing the local stationary solution of the regularized objective.

\paragraph{Efficiency.}
The dynamic pipeline uses one $6.5$\,s acquisition and approximately eight minutes of total wall-clock time. On the same EtH task, EasyHeC takes approximately $32$ minutes, EasyHeC++ $12$ minutes, and our static mode $13$ minutes. Dynamic touch-test error is $0.34$\,cm, compared with $0.28$\,cm for our static mode and $0.30$\,cm for EasyHeC++. The dynamic board-reference translation error is $0.21$\,cm.

\subsection{Multi-Camera and Hardware Transfer}
\label{sec:general}
\begin{figure}[t]
\centering
\includegraphics[width=\linewidth]{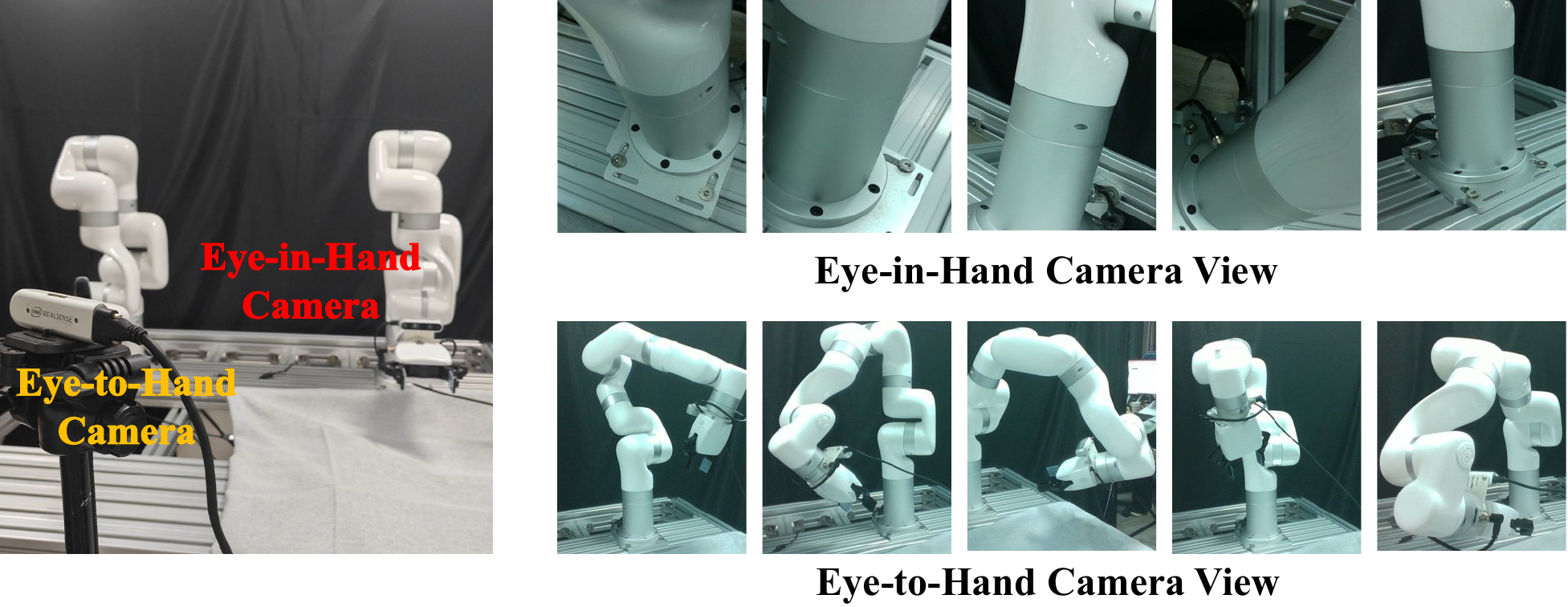}
\caption{Heterogeneous multi-camera co-calibration. \textbf{Left:} the rig, with the tripod-mounted Eye-to-Hand and wrist-mounted Eye-in-Hand cameras labelled. \textbf{Right:} frames from each camera over one continuous trajectory.}
\label{fig:multicam}
\end{figure}

For the dual-camera rig (Figure~\ref{fig:multicam}), independent paired timing measurements give an EtH--EiH offset difference of $0.2$\,ms with a $95\%$ interval of $[-0.4,0.8]$\,ms, contained within the $\pm1$\,ms sharing criterion. This supports the shared-offset variant of Section~\ref{sec:multimethod}. On four matched dual-view recordings, drawn from the five-run benchmark, it obtains $0.45\pm0.27$\,cm / $0.63\pm0.50^\circ$ for EtH and $0.59\pm0.22$\,cm / $1.24\pm0.73^\circ$ for EiH. All four runs converge. The shared fit uses both views in one calibration, while the separate fits attain the single-camera accuracy in Table~\ref{tab:dynamic}.

Table~\ref{tab:general} shows representative runs across seven arms and four depth sensors. The xArm7/D435i configuration is the shared reference, yielding ten distinct hardware configurations. Five independent motions per configuration give $50$ calibration runs. The representative arm runs span $0.24$--$0.68$\,cm translation error and $1.2$--$4.8$\,ms latency error. Across the four sensors, reference offsets range from $66$ to $115$\,ms and estimated offsets lie within $3$\,ms of their references. The more difficult Z1 Pro and Azure Kinect configurations give five-run translation errors of $0.67\pm0.09$ and $0.74\pm0.08$\,cm, respectively. The table's xArm7 entry is a separate EtH run from the five-run mean in Table~\ref{tab:dynamic}.

\begin{table}[t]
\centering
\small
\setlength{\tabcolsep}{3pt}
\caption{Representative runs across arms (top) and depth sensors (bottom). xArm7/D435i is repeated as the shared reference, giving ten distinct configurations. Each configuration has five motions; repeat summaries appear in the text. Offsets are in ms.}
\label{tab:general}
\begin{tabular}{lcccc}
\toprule
\textbf{Platform} & $E_t$ (cm) & $E_r$ ($^\circ$) & $\Delta t^*$ & $\Delta t_{\text{ref}}$\\
\midrule
xArm7 & 0.25 & 0.18 & 85.6 & 87.0\\
Franka Panda & 0.31 & 0.22 & 86.8 & 88.5\\
RM75-B & 0.28 & 0.19 & 86.1 & 87.5\\
Z1 Pro & 0.68 & 0.45 & 84.2 & 89.0\\
EC66 & 0.36 & 0.25 & 86.5 & 88.0\\
ER7 & 0.24 & 0.16 & 87.3 & 88.5\\
CR5 & 0.42 & 0.31 & 85.2 & 87.0\\
\midrule
D435i (Stereo) & 0.25 & 0.18 & 85.6 & 87.0\\
L515 (LiDAR) & 0.18 & 0.13 & 112.4 & 115.0\\
Femto Bolt (ToF) & 0.21 & 0.15 & 64.5 & 66.0\\
Azure Kinect (ToF) & 0.75 & 0.52 & 92.1 & 94.0\\
\bottomrule
\end{tabular}
\end{table}

\begin{table}[t]
\centering
\small
\setlength{\tabcolsep}{3pt}
\caption{Dynamic ablations with the Stage-1 anchor fixed. $E_t$ is in cm, $E_r$ in degrees, and $E_{\rm lat}$ in ms. The full row repeats the five-run means in Table~\ref{tab:dynamic}.}
\label{tab:ablation}
\begin{tabular}{lcccccc}
\toprule
\multirow{2}{*}{\textbf{Variant}} & \multicolumn{3}{c}{\textbf{EtH}} & \multicolumn{3}{c}{\textbf{EiH}}\\
\cmidrule(lr){2-4}\cmidrule(lr){5-7}
& $E_t$ & $E_r$ & $E_{\rm lat}$ & $E_t$ & $E_r$ & $E_{\rm lat}$\\
\midrule
Stage-2 depth off & 1.18 & 0.85 & 3.5 & 1.35 & 0.96 & 8.2\\
Spatial planning only & 0.86 & 0.65 & 16.4 & 0.95 & 0.72 & 18.2\\
\textbf{Full (Ours)} & \textbf{0.21} & \textbf{0.15} & \textbf{2.7} & \textbf{0.24} & \textbf{0.18} & \textbf{5.4}\\
\bottomrule
\end{tabular}
\end{table}

\begin{table}[t]
\centering
\footnotesize
\setlength{\tabcolsep}{2pt}
\caption{Success rate (\%) without $\to$ with recovered-offset compensation. Each cell uses $50$ matched pairs on the six tasks: stack three blocks; place cup, banana, or kettle; open drawer; fold towel. Across $1800$ pairs, the gain is $8.11$ points (paired $95\%$ CI $[6.3,9.9]$).}
\label{tab:downstream}
\begin{tabularx}{\columnwidth}{@{}>{\footnotesize\raggedright\arraybackslash}Xcccccc@{}}
\toprule
\textbf{Policy} & \textbf{Stack} & \textbf{Cup} & \textbf{Ban.} & \textbf{Kettle} & \textbf{Drawer} & \textbf{Towel}\\
\midrule
ACT \citep{zhaoLearningFineGrainedBimanual2023} & $6\!\to\!8$   & $44\!\to\!50$ & $40\!\to\!44$ & $42\!\to\!46$ & $38\!\to\!42$ & $34\!\to\!40$\\
DP \citep{chiDiffusionPolicyVisuomotor2025} & $12\!\to\!16$ & $58\!\to\!64$ & $48\!\to\!56$ & $64\!\to\!68$ & $46\!\to\!54$ & $36\!\to\!52$\\
$\pi_0$ \citep{blackVisionLanguageActionFlow2025} & $16\!\to\!22$ & $54\!\to\!66$ & $50\!\to\!62$ & $46\!\to\!62$ & $48\!\to\!60$ & $62\!\to\!74$\\
DP3 \citep{ze3DDiffusionPolicy2024} & $2\!\to\!4$   & $12\!\to\!16$ & $42\!\to\!48$ & $40\!\to\!48$ & $28\!\to\!32$ & $54\!\to\!66$\\
ManiFlow \citep{yanManiFlowGeneralRobot2025} & $10\!\to\!14$ & $52\!\to\!60$ & $46\!\to\!54$ & $36\!\to\!50$ & $46\!\to\!54$ & $60\!\to\!64$\\
WM$^\dagger$ & $8\!\to\!14$  & $40\!\to\!54$ & $38\!\to\!50$ & $34\!\to\!48$ & $40\!\to\!52$ & $30\!\to\!40$\\
\bottomrule
\end{tabularx}
\par\vspace{2pt}
{\footnotesize\raggedright $^\dagger$WM uses our particle-model MPC implementation based on \citet{hongParticleWorldModel2025} (Appendix~\ref{app:downstream}). All policy results are measured here.\par}
\end{table}

\subsection{Controlled Ablations}
\label{sec:ablations}
\paragraph{Metric depth and completion.}
We hold frames, masks, Stage-1 anchor, and optimizer fixed across three Stage-2 variants. Raw sensor depth gives translation errors of $0.38$/$0.44$\,cm; sensor-aligned completion gives $0.21$/$0.24$\,cm; and disabling the Stage-2 depth term gives $1.18$/$1.35$\,cm (EtH/EiH), averaged over five matched runs. Thus raw metric depth already supplies a useful constraint, and completing missing surface observations further improves pose estimation. Table~\ref{tab:ablation} also shows the associated latency improvement.

\paragraph{Temporal excitation and joint conditioning.}
With $\beta=0$, latency errors increase to $16.4$/$18.2$\,ms. For the stacked seven-column observation Jacobian, after consistent channel scaling and unit-column normalization, the minimum singular values are $0.21$/$0.18$ for the full planner and $0.03$/$0.02$ for spatial-only planning. Both trajectories yield full-rank Jacobians; temporal excitation improves the conditioning of joint pose--time estimation.

\paragraph{Alternation and interpolation.}
Joint Adam with the same per-variable learning rates, batch size, and actual total number of gradient evaluations reaches approximately $0.5$\,cm translation and $6$\,ms latency error over five runs. Alternation reaches the accuracy in Table~\ref{tab:dynamic} in $3$--$5$ cycles. On the same $250$\,Hz joint logs, replacing the linear interpolant with a cubic interpolant changes the estimated translation vector by less than $0.01$\,cm and latency by less than $0.2$\,ms.

\subsection{Downstream Visuomotor Policies}
\label{sec:downstream}
We evaluate six policy families on six real tasks (Table~\ref{tab:downstream}): ACT~\citep{zhaoLearningFineGrainedBimanual2023}, Diffusion Policy (DP)~\citep{chiDiffusionPolicyVisuomotor2025}, $\pi_0$~\citep{blackVisionLanguageActionFlow2025}, DP3~\citep{ze3DDiffusionPolicy2024}, ManiFlow~\citep{yanManiFlowGeneralRobot2025}, and a particle-based world model with sampling-based MPC (WM). Particle dynamics and learned-model control provide the basis for the latter family~\citep{hongParticleWorldModel2025,hafnerDreamControlLearning2020,hafnerMasteringDiverseControl2025,zhuUnifiedWorldModels2025,huangPointWorldScaling3D2026}.

Each task uses the same $50$ demonstrations for both conditions. We fix extrinsics, use one matched training seed per policy--task cell, and replay the same $50$ initial configurations in each pair of evaluations. Without compensation, an image is paired with the nearest joint timestamp. With compensation, it is paired with $\mathbf q(t_c-\dt^*)$, using our recovered offset at both training and deployment. Action and chunk timestamps remain fixed.

Across $36$ cells, $1{,}800$ matched pairs comprise $3{,}600$ physical executions. Successful trials increase from $681$ to $827$. Compensation uniquely succeeds in $210$ pairs and the uncompensated policy uniquely succeeds in $64$, giving $146$ net additional successes and an $8.11$-percentage-point gain. A cell-stratified paired Wald interval is $[6.3,9.9]$ percentage points at $95\%$ confidence, conditional on the trained policies and calibration. Appendix~\ref{app:downstream} gives task definitions, pairing, and the variance calculation.

Mean gains are $4.3$ points for ACT, $7.7$ for DP, $11.7$ for $\pi_0$, $6.0$ for DP3, $7.7$ for ManiFlow, and $11.3$ for WM. Every policy--task cell has a positive observed difference. The aggregate improvement spans image-based, point-cloud, and dynamics-based control, demonstrating the practical value of aligning the observation and joint streams before policy learning and execution.

\subsection{Operating Conditions}
\label{sec:operating}
The pipeline uses a coarse spatial anchor within roughly $15$\,cm, visible robot geometry, and sufficiently excited motion. Strong blur becomes prominent above approximately $0.5$\,m/s in the evaluated imaging setup; acquisition speed should preserve reliable mask boundaries. Pixels without valid depth after completion contribute through the mask channel. A constant offset describes a stable sensing and control stack; changes in the bus, driver, or host state that produce drifts exceeding $20$\,ms call for recalibration. These conditions guide acquisition and deployment.

\section{Conclusion}
SyncHeC combines continuous-time kinematics, metric depth, and planned excitation to jointly estimate hand-eye pose and latency from a single continuous robot motion. It supports EtH, EiH, and heterogeneous hardware. Applying the recovered offset during policy training and deployment raises real manipulation success by $8.1$ points across six families.

\FloatBarrier
\bigskip
\appendix
\section{Optimization and Implementation}
\label{app:optimizer}
\subsection{Update Accounting and Stopping}
Algorithm~\ref{alg:alt} specifies the alternating optimizer. A pass visits every mini-batch in the active frame set once. With $N=60$, batch size $B=4$, and $P=15$ passes, each block executes
\begin{equation}
P\lceil N/B\rceil=15\times15=225
\end{equation}
Adam updates. There is one optimizer step per batch and no gradient accumulation. The temporal gate is evaluated at the current synchronized times at the start of each cycle; all $60$ selected benchmark frames pass it. Each cycle therefore uses $225$ temporal and $225$ spatial gradient evaluations. Three to five cycles correspond to $675$--$1{,}125$ temporal updates and $1{,}350$--$2{,}250$ total gradient evaluations; the maximum of $15$ cycles allows $3{,}375$ temporal updates.

For a mini-batch $\mathcal B$, the optimized objective is the mean frame data loss plus one regularization term:
\begin{equation}
\mathcal L_{\mathcal B}
=\frac{1}{|\mathcal B|}\sum_{i\in\mathcal B}
  (\mathcal L_{m,i}+\lambda_D\mathcal L_{d,i})
  +\lambda_{\rm reg}\dt^2.
\label{eq:batchloss}
\end{equation}
This keeps the penalty's scale consistent with Equation~\eqref{eq:loss}. Time is stored in seconds, translation in meters, and rotation updates use the Lie-algebra representation. Separate optimizers retain their moments across passes and alternating cycles.

\begin{algorithm}[htbp]
\caption{Alternating pose and latency optimization}
\label{alg:alt}
\small
\textbf{Input:} selected frames, joint log, mesh, intrinsics, $\mathbf T^0$, $P=15$, $B=4$, $K_{\max}=15$.
\begin{enumerate}
\setlength{\itemsep}{3pt}
\item Initialize $\xi=\log\mathbf T^0$, $\dt=0$\,s, and the spatial and temporal Adam states.
\item For each cycle $k=1,\ldots,K_{\max}$:
\begin{enumerate}
\setlength{\itemsep}{3pt}
\item Save the current extrinsic and offset for the stopping test. Select $\mathcal S=\{i:\|\dot{\mathbf q}(t_i^c-\dt)\|_2>0.05\ {\rm rad/s}\}$.
\item Fix $\xi$. For each of $P$ passes over $\mathcal S$, form batches of $B$ frames, recompute active interpolation intervals, evaluate $\mathcal L_{\mathcal B}$, and take one Adam update of $\dt$ per batch.
\item Fix $\dt$. Resynchronize all $N$ frames. For each of $P$ passes, form batches of $B$ frames and take one Adam update of $\xi$ per batch.
\item Stop if all cycle-to-cycle changes satisfy: translation $<0.05$\,mm, rotation $<0.01^\circ$, and time $<0.5$\,ms.
\end{enumerate}
\item Return $\mathbf T^*=\exp(\xi)$ and $\dt^*$.
\end{enumerate}
\end{algorithm}

\subsection{Representative Trace and Regularization}
Table~\ref{tab:trace} records one EtH run from the dynamic benchmark. The pose is initialized by Stage~1 and the temporal variable starts at zero. Its final offset, $84.3$\,ms, is also close to the five-run mean.

\begin{table}[t]
\centering
\small
\caption{Representative EtH optimization trace. The count is the cumulative number of temporal optimizer updates.}
\label{tab:trace}
\begin{tabular}{lrrrrr}
\toprule
Temporal updates & 0 & 225 & 450 & 675 & 900\\
Offset (ms) & 0 & 43.8 & 78.1 & 84.0 & 84.3\\
\bottomrule
\end{tabular}
\end{table}

Writing the mean data objective as $\mathcal L_{\rm data}$, the temporal stationarity condition is
\begin{equation}
\frac{\partial\mathcal L_{\rm data}}{\partial\dt}
 +2\lambda_{\rm reg}\dt=0.
\label{eq:stationarity}
\end{equation}
At $\dt=0.0843$\,s and $\lambda_{\rm reg}=0.5$, the penalty gradient is $+0.0843$\,s$^{-1}$, balancing the observed data gradient of approximately $-0.0843$\,s$^{-1}$. Retaining optimizer states and continuing the original alternating sequence for $900$ further temporal updates, together with the associated spatial blocks, changes the final offset by less than $0.2$\,ms with stopping disabled. The penalty's influence is governed by the local data gradient and curvature.

\subsection{Software Implementation}
\label{app:implementation}
SyncHeC offers Nvdiffrast rendering on CUDA and soft rendering in PyTorch. Both use the same continuous-time kinematics, per-frame normalized losses, and mini-batch alternating optimizer. Configuration files specify batch size, passes per block, loss weights, and learning rates; each run records the selected frames, recovered transform, offset, and actual update counts.

Recordings supply camera intrinsics, time-stamped joints, robot masks, metric depth, and URDF geometry. The depth-fusion function preserves valid sensor measurements and completes holes using sensor-scaled predictions. Spatial initialization uses a separate base-link mask and the surface-to-center correction. EtH and EiH use the transforms in Equations~\eqref{eq:eth}--\eqref{eq:eih}; multi-camera calibration supports camera-specific or shared offsets.

The acquisition pipeline uses rendered-pixel trajectory scoring with RRT* and collision checks. The software also provides a lightweight B-spline refinement interface based on projected URDF landmarks, with collision and visibility callbacks. Table~\ref{tab:code} maps the calibration components to their implementation.

\begin{table}[t]
\centering
\footnotesize
\caption{Core mathematical components in SyncHeC.}
\label{tab:code}
\begin{tabular}{lp{0.49\linewidth}}
\toprule
Component & Function\\
\midrule
Calibration objective & \texttt{optimize.CalibrationProblem}\\
Joint interpolation & \texttt{kinematics.q\_sync}\\
Temporal derivative & \texttt{kinematics.temporal\_jacobian}\\
EtH / EiH transforms & \texttt{kinematics.LinkPoseModel}\\
Mask / depth losses & \texttt{render.losses}\\
Initialization fitness & \texttt{stage1.joint\_fitness}\\
Alternating optimizer & \texttt{optimize.calibrate}\\
Joint optimizer & \texttt{optimize.calibrate\_joint}\\
\bottomrule
\end{tabular}
\end{table}

The executable synthetic example generates a four-link robot recording with a known $87$\,ms offset, initializes pose by perturbing the reference extrinsic, and starts the temporal variable at zero. Its default configuration uses $24$ frames at $64\times48$ resolution, $\lambda_{\rm reg}=0$, temporal learning rate $0.001$, and five passes over batches of four frames. Each block therefore performs $30$ updates per cycle. The output reports pose and timing errors against the generating parameters together with the complete optimization history.

\section{Metric Depth and Anchoring}
\label{app:depth}
\subsection{Metric Fusion and Validity}
Let $\mathcal V_s$ contain valid raw sensor pixels and $\mathcal V_p$ contain finite, positive Depth Anything V2 predictions. Using the ViT-L metric Hypersim checkpoint, we compute
\begin{equation}
s=\operatorname{median}_{p\in\mathcal V_s\cap\mathcal V_p}
\frac{D_{\rm raw}^{(p)}}{D_{\rm pred}^{(p)}}.
\end{equation}
With overlap available for this scale estimate, the observation depth is
\begin{equation}
D_{\rm obs}^{(p)}=
\begin{cases}
D_{\rm raw}^{(p)}, &p\in\mathcal V_s,\\
sD_{\rm pred}^{(p)}, &p\in\mathcal V_p\setminus\mathcal V_s.
\end{cases}
\end{equation}
Pixels without a valid retained measurement or scaled prediction remain invalid. The per-frame depth-supervision region is
\begin{equation}
\Omega_i=\{p:M_i^{{\rm rend},(p)}>0.5,\ 
M_i^{{\rm obs},(p)}=1,\ p\in\mathcal V_{{\rm obs},i}\}.
\end{equation}
Mask residuals are evaluated across the full image. The scale is anchored to sensor measurements, and valid sensor depth is preserved throughout the fusion.

The raw-depth control uses $\mathcal V_s$ for its Stage-2 depth residual. The completed-depth control uses $\mathcal V_{\rm obs}$. The $\lambda_D=0$ control switches off this Stage-2 residual. All three use the same completed-depth Stage-1 anchor, frames, masks, and optimizer, isolating the contribution of Stage-2 depth supervision and completion.

\subsection{Surface-to-Base Correction}
The median depth over valid segmented base-link pixels yields $d_{\rm surf}$. Under the base-centered viewing approximation used for candidate sampling, this gives a surface-distance estimate. The side-profile model relates this estimate, camera height $H$ relative to the base origin, and base cylinder radius $r_{\rm base}$ to center distance $L$:
\begin{equation}
L=d_{\rm surf}+
\frac{r_{\rm base}L}{\sqrt{L^2-H^2}}.
\label{eq:side}
\end{equation}
It describes the visible cylindrical side in the regime $d_{\rm surf}>|H|$ and $L\gg r_{\rm base}$. A first-order correction evaluates the right-hand side at $L_0=d_{\rm surf}$, then sets the sampling radius to $L+\Delta r$. The joint mask--depth fitness scores these approximate candidates before rendering refinement. The dynamic pipeline uses one static frame for this anchor; the static benchmark adds refinement over five views.

\section{Excitation and Joint Conditioning}
\label{app:conditioning}
\subsection{Sensitivity Interpretation}
For a differentiable observation vector $\mathbf y(\theta)$ with independent Gaussian noise of variance $\sigma^2$, the Fisher matrix is
\begin{equation}
\mathbf F=\frac{1}{\sigma^2}\mathbf J^\top\mathbf J,
\qquad
\mathbf J=\frac{\partial\mathbf y}{\partial\theta}.
\end{equation}
For small camera-pose perturbations with covariance $\mathbf C_\xi$, linearization gives
\begin{equation}
V_{\rm space}\approx
\operatorname{tr}(\mathbf J_\xi\mathbf C_\xi\mathbf J_\xi^\top).
\end{equation}
For latency, the corresponding per-frame information is
\begin{equation}
F_{tt}=\frac{\|\mathbf J_\pi\dot{\mathbf q}\|_F^2}{\sigma^2}
=\frac{V_{\rm time}^2}{\sigma^2}.
\end{equation}
The planner uses $\sum_kV_{\rm time}$ alongside the spatial variance surrogate. The unsquared temporal form reduces the relative reward for isolated high-amplitude peaks compared with $\sum_kV_{\rm time}^2$. Fixed reference scales make the relative weight $\gamma$ dimensionless and comparable across planning iterations. The joint Jacobian diagnostic below evaluates pose--time coupling after planning.

\subsection{Seven-Parameter Diagnostic}
For each EtH or EiH sequence, we stack the six pose derivatives and one temporal derivative of the mask and valid-depth observation channels. The diagnostic uses consistent channel weights, pixel-count normalization, and sampling rules across variants. It differentiates the data observations, excluding the latency penalty. Let $\mathbf J_w$ be the scaled matrix and
\begin{equation}
\widetilde{\mathbf J}
=\mathbf J_w\operatorname{diag}
\left(\|\mathbf J_{w,:,1}\|_2^{-1},\ldots,
\|\mathbf J_{w,:,7}\|_2^{-1}\right).
\end{equation}
Unit-column normalization makes the singular-value comparison independent of the units used for translation, rotation, and time.

\begin{table}[t]
\centering
\small
\caption{Minimum singular value of the normalized, stacked data Jacobian.}
\label{tab:conditioning}
\begin{tabular}{lcc}
\toprule
Trajectory & EtH & EiH\\
\midrule
Spatial excitation only ($\beta=0$) & 0.03 & 0.02\\
Spatial and temporal excitation & 0.21 & 0.18\\
\bottomrule
\end{tabular}
\end{table}

All four matrices in Table~\ref{tab:conditioning} are full rank. Because seven columns have unit norm, $\|\widetilde{\mathbf J}\|_F^2=7$ and $1\leq\sigma_{\max}\leq\sqrt7$. The full planner therefore has condition number at most $12.6$/$14.7$ (EtH/EiH); spatial-only planning has condition number at least $33.3$/$50$. These bounds quantify improved conditioning for the joint estimation problem.

Structural first-order ambiguity occurs when one fixed pose perturbation reproduces the effect of a time shift across all stacked observations. A constant rigid-body twist can produce such a case. Constant joint velocity alone does not determine the stacked column relationship, since articulated kinematics, perspective, and visibility also change along a trajectory.

\section{Timing Reference and Comparison Protocols}
\label{app:timing}
\subsection{Reference Error Budget}
The reference maps image exposure and physical motion onset into a common timing measurement. The recorded event is the robot's actual motion onset in the joint stream. Exposure-phase measurements include the RGB--depth timestamp relation, and clock mapping accounts for the two stream origins. Table~\ref{tab:timingbudget} lists the residual error budget.

\begin{table}[t]
\centering
\small
\caption{Uncertainty budget for the nominal $87$\,ms reference. The first three entries are residual bias bounds; the fourth is a confidence half-width for the repeated mean.}
\label{tab:timingbudget}
\begin{tabular}{lr}
\toprule
Component & ms\\
\midrule
Exposure phase, including RGB--depth skew & 0.25\\
Clock mapping & 0.10\\
Physical motion-onset localization & 0.20\\
Repeated-mean $95\%$ half-width & 0.95\\
\midrule
Combined $95\%$ uncertainty $U_{95}$ & 1.50\\
\bottomrule
\end{tabular}
\end{table}

The $20$ repetitions characterize repeatability; the exposure-phase and onset measurements supply sub-frame timing. Adding the three residual bias bounds to the repeated-mean interval gives $U_{95}=0.25+0.10+0.20+0.95=1.50$\,ms.

For a reference adjustment $b\in[-1.5,1.5]$\,ms, the EtH mean offsets give
\begin{equation}
R(b)=
\frac{|87+b-80.8|}{|87+b-84.3|}
=\frac{6.2+b}{2.7+b}
\in[1.83,3.92].
\label{eq:ratio}
\end{equation}
Thus the ordering is maintained across the reference interval. The EiH means, $78.5$ and $81.6$\,ms, likewise retain their ordering. Spatial evaluation uses separate marker-board and touch references.

\subsection{Paired Multi-Camera Timing}
The independent timing reference gives an EtH--EiH difference of $0.2$\,ms and a $95\%$ interval $[-0.4,0.8]$\,ms. Its containment in $[-1,1]$\,ms motivates the shared-offset model for the four paired dual-view recordings. Common reference errors cancel in this difference, giving a tighter differential interval than the absolute reference budget. Camera-specific regularized calibration estimates ($84.3$ and $81.6$\,ms) reflect each view's data objective as well as the penalty; the sharing criterion is evaluated using the independent timing reference.

\subsection{Matched Baselines and Controls}
\label{app:comparisons}
\paragraph{Static protocol.}
The marker-based, DREAM, EasyHeC, EasyHeC++, Zhang et al., and Valassakis et al.\ entries in Tables~\ref{tab:sim}--\ref{tab:real_static} are taken from the corresponding EasyHeC++ comparison~\citep{hongEasyHeCFullyAutomatic2024}. Our static reruns use the same simulation/touch evaluation definitions, and the Hydra comparison uses the matched real static protocol. The touch evaluation averages ten poses.

\paragraph{Dynamic protocol.}
CM-Calib, SL-Calib, and SyncHeC share recordings, selected trajectory, and joint logs; initialization is matched where applicable. CM-Calib uses target-derived poses and forward kinematics. The fixed-reference control sets $\dt=87$\,ms for EasyHeC++ and the SyncHeC spatial objective while preserving their remaining inputs, initialization, and pose-update rules. This yields $0.43$/$0.48$\,cm and $0.20$/$0.23$\,cm, respectively.

\paragraph{Optimization control.}
Joint Adam and alternating Adam use the same per-variable learning rates and batch size, and are compared at the same actual total count of data-gradient evaluations. A joint update evaluates both variable gradients in one forward/backward computation; an alternating update evaluates the active block. The evaluation budget counts executed computations, including the treatment of early stopping.

\paragraph{Hardware repetitions.}
Seven arm configurations and four sensor configurations share the xArm7/D435i reference, giving ten unique configurations. Five motions per configuration give $50$ runs in total, including the ten representative runs in Table~\ref{tab:general}. The additional repetitions supply the Z1 Pro and Azure Kinect means and standard deviations reported in Section~\ref{sec:general}.

\section{Downstream Study: Protocol and Statistics}
\label{app:downstream}
\begin{table*}[t]
\centering
\small
\setlength{\tabcolsep}{4pt}
\caption{The six real-robot tasks: regime, randomization of the initial object
configuration, and success criterion. Yaw ranges are about the vertical axis.}
\label{tab:supp_tasks}
\begin{tabular}{llp{175pt}p{130pt}}
\toprule
\textbf{Task} & \textbf{Regime} & \textbf{Randomization of the initial configuration} & \textbf{Success criterion}\\
\midrule
Stack three blocks & long-horizon & three $20{\times}40$\,cm cells, $\pm5$\,cm jitter
  & all three stacked, tower stable at timeout\\
Place cup & pose-constrained & $60{\times}40$\,cm board, $360^\circ$ yaw
  & cup upright inside the target region\\
Place banana on plate & pose-constrained & plate on midline $+10$\,cm; banana
  $10$--$15$\,cm left, $360^\circ$ yaw & banana resting on the plate\\
Place kettle on stove & pose-constrained & $20{\times}20$\,cm region, $\pm60^\circ$
  & kettle upright on the stove plate\\
Open drawer & articulated & L-shaped region ($70{\times}20$ $+$ $20{\times}20$\,cm),
  rotated $90^\circ$ & drawer pulled to $\geq50\%$ extension\\
Fold towel & deformable & $70{\times}50$\,cm region, $\pm30^\circ$
  & towel folded, free edge within tolerance\\
\bottomrule
\end{tabular}
\end{table*}

\subsection{Controlled Pairing}
The study varies image--joint temporal pairing while fixing robot hardware, cameras, extrinsics, demonstrations, architecture, training hyperparameters, and the matched training seed within each policy--task cell. Each task provides $50$ successful teleoperated demonstrations. The same recordings train both conditions:
\begin{itemize}
\setlength{\itemsep}{2pt}
\item \textbf{Uncompensated:} pair each image with the joint state carrying the nearest recorded timestamp.
\item \textbf{Compensated:} pair each image at $t_c$ with the interpolated joint state at $t_c-\dt^*$.
\end{itemize}
The same rule applies during deployment. Compensation uses the recovered calibration estimate. Action timestamps and action-chunk scheduling are fixed across conditions.

\subsection{Observations and Policy Families}
The robot uses the two D435i cameras at $30$\,Hz and joint logs at $250$\,Hz. The constant offset includes the stream-clock difference and sensing-to-availability delay. For point-cloud policies, offline demonstration processing enhances depth, transforms both views into the robot base frame, merges and crops them to the workspace, and applies farthest-point sampling to $8192$ points. At deployment, point clouds are produced online at approximately $30$\,Hz with offline depth enhancement disabled. Both temporal conditions use this same observation pipeline.

ACT uses RGB observations and an action-chunking transformer~\citep{zhaoLearningFineGrainedBimanual2023}; DP uses RGB-conditioned action diffusion~\citep{chiDiffusionPolicyVisuomotor2025}; $\pi_0$ is fine-tuned from its released vision-language-action checkpoint~\citep{blackVisionLanguageActionFlow2025}. DP3~\citep{ze3DDiffusionPolicy2024} and ManiFlow~\citep{yanManiFlowGeneralRobot2025} use the point clouds above. The WM family models particle-based 3D dynamics and uses MPPI-style sampling-based MPC with $1000$ action samples per iteration and $15$--$30$ refinement iterations. Particle-based dynamics learned from human interaction and model-based control motivate this representation~\citep{hongParticleWorldModel2025,zhuUnifiedWorldModels2025,hafnerMasteringDiverseControl2025}.

\subsection{Tasks and Execution}
Table~\ref{tab:supp_tasks} defines the six tasks. For each of the $36$ policy--task cells, $50$ independently randomized initial configurations are replayed in both conditions. Initial-state randomization is independent across cells. A trial succeeds if its task criterion is met within $60$\,s; timeouts, dropped objects, and unrecoverable states count as failures. Each cell therefore contains $50$ matched pairs and $100$ physical executions.

\subsection{Paired Aggregate Effect}
\label{app:paired}
Let $X_{ji}\in\{-1,0,1\}$ denote the difference in binary success between conditions for pair $i$ in cell $j$. With $n_j=50$ and $N=\sum_jn_j=1800$, the aggregate estimate is
\begin{equation}
\widehat\delta=\frac1N\sum_{j=1}^{36}\sum_{i=1}^{n_j}X_{ji}.
\end{equation}
For each cell, let $b_j$ and $c_j$ count pairs won only with and only without compensation. Then $d_j=b_j-c_j$ is fixed by the cell's marginal success counts, and the sample variance is
\begin{equation}
s_j^2=
\frac{b_j+c_j-d_j^2/n_j}{n_j-1}.
\end{equation}
The variance of the stratified aggregate, conditional on the trained policies and calibration, is estimated by
\begin{equation}
\widehat{\operatorname{Var}}(\widehat\delta)
=\frac1{N^2}\sum_{j=1}^{36}n_js_j^2.
\label{eq:pairedvariance}
\end{equation}
With equal cell sizes, this reduces to
\begin{equation}
\widehat{\operatorname{Var}}(\widehat\delta)
=\frac{50(b+c)-\sum_jd_j^2}{49N^2}.
\end{equation}
The counts satisfy $b=210$, $c=64$, and $\sum_jd_j^2=740$. The estimated gain is $\widehat\delta=146/1800=0.08111$ and its standard error is $0.00904$. The paired Wald interval $\widehat\delta\pm1.96\,\mathrm{SE}$ is $[0.0634,0.0988]$, or $[6.3,9.9]$ percentage points.

\begin{table}[htbp]
\centering
\small
\caption{Aggregate paired outcomes over $1800$ initial configurations.}
\label{tab:paired}
\begin{tabular}{lrr}
\toprule
& \multicolumn{2}{c}{Compensated}\\
\cmidrule(lr){2-3}
Uncompensated & Success & Failure\\
\midrule
Success & 617 & 64\\
Failure & 210 & 909\\
\bottomrule
\end{tabular}
\end{table}

This interval measures execution variability for the fixed trained policies and calibration. The $36$ cell differences and six family averages describe how the aggregate gain is distributed across tasks and architectures. The total counts in Table~\ref{tab:paired} agree with the $681$ and $827$ successes obtained from Table~\ref{tab:downstream}.

\clearpage
\begingroup
\small
\bibliography{references}
\endgroup
\end{document}